\RequirePackage{fix-cm}
\documentclass[10pt]{article}

\usepackage{etoolbox}
\newcommand{\arxiv}[1]{\iftoggle{iclr}{}{#1}}
\newcommand{\iclr}[1]{\iftoggle{iclr}{#1}{}}
\newtoggle{iclr}
\global\togglefalse{iclr}

\PassOptionsToPackage{dvipsnames}{xcolor}

\usepackage[utf8]{inputenc}
\usepackage[T1]{fontenc}
\usepackage{url}
\usepackage{booktabs}
\usepackage{amsfonts}
\usepackage{nicefrac}
\usepackage{microtype}
\usepackage{bm}
\usepackage{dsfont}
\usepackage{wrapfig}
\usepackage{enumitem}

\usepackage{colortbl}
\usepackage{xcolor}

\usepackage{inconsolata}
\DeclareTextCommandDefault{\textasciigrave}{\UseTextSymbol{TS1}\textasciigrave}
\usepackage[scaled=.90]{helvet}
\usepackage{xspace}
\usepackage[most]{tcolorbox}
\usepackage{pifont}

\usepackage{amsmath,amsfonts,bm}

\def\eqref#1{equation~\ref{#1}}

\def\1{\bm{1}}

\DeclareMathAlphabet{\mathsfit}{\encodingdefault}{\sfdefault}{m}{sl}
\SetMathAlphabet{\mathsfit}{bold}{\encodingdefault}{\sfdefault}{bx}{n}

\arxiv{
\usepackage[a4paper, margin=2.5cm, headheight=28.5pt, headsep=3mm]{geometry}
\PassOptionsToPackage{hypertexnames=false}{hyperref}
\usepackage[colorlinks,hyperfootnotes=false]{hyperref}
\hypersetup{
    citecolor=[RGB]{50,100,170},
    linkcolor=[RGB]{50,100,170},
    urlcolor=[RGB]{255,102,178}}
\usepackage{fancyhdr}
}

\arxiv{
\newcommand{\toptitlebar}{%
  {\color{black}\hrule height 1pt}%
  \vskip 0.25in%
}
}

\makeatletter
\arxiv{
\renewcommand{\maketitle}{%
  \begin{center}%
    \toptitlebar
    \vskip 0.1in%
    {\LARGE\bfseries \@title \par}%
    \vskip 0.3in%
    {\normalsize \@author \par}%
  \end{center}%
  \par
  \vskip 0.3in%
}

\renewcommand\section{\@startsection {section}{1}{\z@}{-2.0ex plus
    -0.5ex minus -.2ex}{1.5ex plus 0.3ex minus .2ex}{\large\bfseries\raggedright}}
\renewcommand\subsection{\@startsection{subsection}{2}{\z@}{-1.8ex plus
    -0.5ex minus -.2ex}{0.8ex plus .2ex}{\normalsize\bfseries\raggedright}}
\renewcommand\subsubsection{\@startsection{subsubsection}{3}{\z@}{-1.5ex plus
   -0.5ex minus -.2ex}{0.5ex plus .2ex}{\normalsize\bfseries\raggedright}}

\renewenvironment{abstract}%
  {\centerline{\large\bfseries Abstract}%
   \begin{list}{}%
      {\setlength{\rightmargin}{0.6cm}%
       \setlength{\leftmargin}{0.6cm}}%
    \item[]\ignorespaces}%
  {\unskip\end{list}}

}
\makeatother

\iclr{
\PassOptionsToPackage{hypertexnames=false}{hyperref}
\usepackage[colorlinks=true,breaklinks=true]{hyperref}
}

\usepackage{hhline}

\makeatletter
\newcommand{\neutralize}[1]{\expandafter\let\csname c@#1\endcsname\count@}
\makeatother

\usepackage{algorithm}

\arxiv{
\usepackage{natbib}
\bibliographystyle{plainnat}
\bibpunct{(}{)}{;}{a}{,}{,}
}

\usepackage{amsmath}
\usepackage{mathtools}
\usepackage{amsthm}
\usepackage{amssymb}

\usepackage{xpatch}

\usepackage{thmtools}
\makeatletter
\@ifundefined{newcounteralias}{}{%
  \renewcommand\thmt@autorefsetup{%
    \expandafter\def\csname\thmt@envname autorefname\expandafter\endcsname
      \expandafter{\thmt@thmname}}}
\makeatother
\declaretheorem[name=Theorem]{theorem}
\declaretheorem[name=Lemma,sibling=theorem]{lemma}
\declaretheorem[name=Assumption,sibling=theorem]{assumption}
\declaretheorem[name=Condition,sibling=theorem]{condition}

\declaretheorem[name=Proposition,sibling=theorem]{proposition}

\makeatletter
  \renewenvironment{proof}[1][Proof]%
  {%
   \pushQED{\qed}%
   \par\noindent{\bfseries\upshape {#1.}\ }%
  }%
  {\popQED\par}
  \makeatother

\theoremstyle{definition}

\theoremstyle{plain}
\newtheorem{corollary}[theorem]{Corollary}
\newtheorem{definition}{Definition}[section]

\usepackage[nameinlink,capitalize]{cleveref}

\crefformat{equation}{#2Eq.~(#1)#3}
\Crefformat{equation}{#2Eq.~(#1)#3}

\Crefformat{figure}{#2Figure~#1#3}
\Crefname{assumption}{Assumption}{Assumptions}
\Crefformat{assumption}{#2Assumption~#1#3}
\Crefname{subsubsection}{Section}{Sections}
\crefformat{subsubsection}{#2Section~#1#3}
\Crefformat{subsubsection}{#2Section~#1#3}
\crefname{algorithm}{Alg.}{Algs.}
\Crefname{algorithm}{Alg.}{Algs.}

\usepackage{crossreftools}
\usepackage{xparse}

\ExplSyntaxOn
\DeclareDocumentCommand{\XDeclarePairedDelimiter}{mm}
 {
  \__egreg_delimiter_clear_keys:
  \keys_set:nn { egreg/delimiters } { #2 }
  \use:x
   {
    \exp_not:n {\NewDocumentCommand{#1}{sO{}m} }
     {
      \exp_not:n { \IfBooleanTF{##1} }
       {
        \exp_not:N \egreg_paired_delimiter_expand:nnnn
         { \exp_not:V \l_egreg_delimiter_left_tl }
         { \exp_not:V \l_egreg_delimiter_right_tl }
         { \exp_not:n { ##3 } }
         { \exp_not:V \l_egreg_delimiter_subscript_tl }
       }
       {
        \exp_not:N \egreg_paired_delimiter_fixed:nnnnn 
         { \exp_not:n { ##2 } }
         { \exp_not:V \l_egreg_delimiter_left_tl }
         { \exp_not:V \l_egreg_delimiter_right_tl }
         { \exp_not:n { ##3 } }
         { \exp_not:V \l_egreg_delimiter_subscript_tl }
       }
     }
   }
 }

\keys_define:nn { egreg/delimiters }
 {
  left      .tl_set:N = \l_egreg_delimiter_left_tl,
  right     .tl_set:N = \l_egreg_delimiter_right_tl,
  subscript .tl_set:N = \l_egreg_delimiter_subscript_tl,
 }

\cs_new_protected:Npn \__egreg_delimiter_clear_keys:
 {
  \keys_set:nn { egreg/delimiters } { left=.,right=.,subscript={} }
 }

\cs_new_protected:Npn \egreg_paired_delimiter_expand:nnnn #1 #2 #3 #4
 {%
  \mathopen{}
  \mathclose\c_group_begin_token
   \left#1
   #3
   \group_insert_after:N \c_group_end_token
   \right#2
   \tl_if_empty:nF {#4} { \c_math_subscript_token {#4} }
 }
\cs_new_protected:Npn \egreg_paired_delimiter_fixed:nnnnn #1 #2 #3 #4 #5
 {
  \mathopen{#1#2}#4\mathclose{#1#3}
  \tl_if_empty:nF {#5} { \c_math_subscript_token {#5} }
 }
\ExplSyntaxOff

\XDeclarePairedDelimiter{\supnorm}{
  left=\lVert,
  right=\rVert,
  subscript=\infty
  }

\definecolor{citationcolor}{RGB}{50,100,170}
\definecolor{urlcolor}{RGB}{255,102,178}
\definecolor{postcutoff}{RGB}{236,235,243}
\definecolor{humanrow}{RGB}{229,242,232}
\definecolor{schemarow}{RGB}{255,232,228}
\definecolor{opgreen}{HTML}{006E45}
\definecolor{opblue}{HTML}{1A79EF}
\definecolor{opslate}{HTML}{63778F}

\usepackage{graphicx}

\usepackage{etoc}

\usepackage{caption}
\usepackage{parskip}
\DeclareHookRule{begindocument}{parskip}{before}{hyperref}
\usepackage{needspace}
\newcommand{\wrapneedspace}[1]{\Needspace{#1\baselineskip}}

\usepackage{subcaption}
\usepackage{array}
\usepackage{multirow}
\usepackage{siunitx}
\usepackage{xurl}
\usepackage{algorithmic}

\definecolor{dicegreen}{HTML}{5F7A3B}
\newcommand{\dicestage}[1]{\item[]\textcolor{dicegreen}{$\triangleright$\,\textit{#1}}}

\title{{\fontsize{15.5}{19}\selectfont Speculative Decoding Is Less Speculative Than You Think}}
\author{
  \textbf{Guanning Zeng} \quad
  \textbf{Andrea Zanette} \\[2mm]
  {\small Carnegie Mellon University}
}

\hypersetup{
  pdftitle={Speculative Decoding Is Less Speculative Than You Think},
  pdfauthor={Guanning Zeng and Andrea Zanette},
}

\begin{document}
\etocdepthtag.toc{maintext}

\maketitle

\thispagestyle{fancy}
\fancyhead{}
\lhead{\raisebox{-0.66cm}{\includegraphics[height=0.55cm]{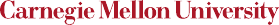}}}
\renewcommand{\headrulewidth}{0pt}

\begin{abstract}
Speculative decoding accelerates language-model inference by preparing candidate continuations before the target model generates them. Its efficiency depends on how well limited drafting computation anticipates the path the target will actually follow. Anticipating this path involves two sources of uncertainty: the target model's predictions and the randomness of sampling. Yet the former requires a target-model forward pass, while the latter can be fixed before drafting. Building on this observation, we introduce \textbf{DICE} (Draw-Informed Candidate Selection), which pre-samples and synchronizes verification randomness across drafting and verification. By combining these known draws with predictions from an existing drafter, DICE allocates computation toward continuations more likely to be used in the current sampling run. This principle guides node selection throughout draft-tree construction and extends to asynchronous prefetching, where DICE predicts which next-round drafts to prepare. Without additional drafter training or changes to the target sampling distribution, DICE achieves state-of-the-art decoding throughput across diverse settings, with speedups of up to 24.6\% over strong baselines.
\end{abstract}

\section{Introduction}

Speculative decoding (SD) is the art of anticipating the future.
Before a large target model generates its next tokens, a lightweight
drafter prepares plausible continuations, allowing the target to
replace several sequential inference steps with a single parallel
verification pass.
But the drafter cannot afford to explore every possible continuation. 
Its efficiency therefore depends on allocating limited computation to the paths the target will actually follow.

Anticipating the target's continuation involves two sources of
uncertainty: \emph{unknown target predictions} and
\emph{sampling randomness}.
The former arises because the drafter cannot fully predict
the target model's output distribution.
The latter persists even with fixed target and draft
distributions, as the sampling outcome can vary with
the random draws.
For example, in standard speculative sampling
\citep{leviathan2023,chen2023}, a proposed token $x$ is accepted if
\begin{equation}
    u\,q(x) \leq p(x),
    \label{eq:acceptance}
\end{equation}
where $p(x)$ and $q(x)$ are its target and draft probabilities,
respectively, and $u\sim\mathrm{Uniform}(0,1)$ is the
verification draw.
Much of the progress in speculative decoding has focused
on the former, improving draft architectures
\citep{cai2024medusa,li2024eagle,chen2026dflash}
and training methods
\citep{zhou2024distillspec,zhang2025hass,li2025eagle3}
to bring the draft distribution $q$ closer to the target $p$.

Yet the two sources of uncertainty differ in one crucial respect:
obtaining the target's predictions requires a costly forward pass,
while the random draws used for verification can be generated
before drafting begins.
Could this very source of uncertainty help us decide
what to draft?

\begin{wrapfigure}{r}{0.44\linewidth}
  \vspace{-10pt}
  \centering
  \includegraphics[width=\linewidth]{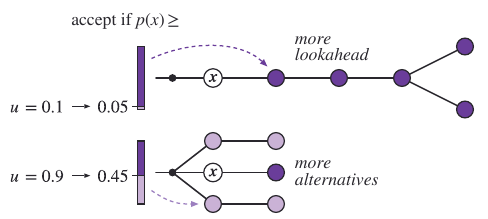}
  \caption{\textbf{Known sampling randomness guides speculative computation.}
  Both rows share the same proposal $x$ with $q(x)=0.5$,
  but differ in the verification draw $u$.
  Dark and light purple indicate the values of $p(x)$ that would
  accept or reject $x$, respectively.
  The dashed arrows illustrate how the same drafting budget
  can be allocated toward more lookahead through $x$ or
  more alternatives in case of rejection.}
  \label{fig:intro}
  \vspace{-6pt}
\end{wrapfigure}

Figure~\ref{fig:intro} illustrates this with a fixed
proposal $x$ and $q(x)=0.5$.
When $u=0.1$, $x$ is accepted whenever
$p(x)\geq 0.05$; when $u=0.9$, acceptance requires
$p(x)\geq 0.45$.
\textbf{The target probability is still unknown, but the
acceptance threshold is already known!}
A small draw may favor looking further ahead through
$x$, while a large draw may favor preparing alternatives
in case $x$ is rejected.
Thus, the same draft prediction can lead to different
allocations of speculative computation.

\wrapneedspace{16}
The same idea also applies to Gumbel-max sampling, where
the random perturbations used for target sampling can
likewise be generated before drafting.
Building on this observation, we introduce \textbf{DICE}
(Draw-Informed Candidate Selection),
named for the random draws that guide its drafting decisions.
DICE combines these draws with draft-model predictions
to guide speculative computation.
In draft-tree construction, the resulting scores guide
candidate generation, tree expansion, and final node selection.
We further apply DICE to asynchronous prefetching,
using known random draws to predict verification outcomes
and prioritize next-round drafts.
DICE works with existing draft models and preserves
the exact target sampling distribution.

We evaluate DICE across diverse target--drafter pairs,
generation tasks, and inference settings.
Under matched tree-node and prefetch-cache budgets,
DICE improves acceptance length and cache hit rate,
yielding decoding throughput gains of up to 24.6\%
and 13.7\% in tree-based decoding and asynchronous
prefetching, respectively.
In optimized inference engines such as SGLang and vLLM,
DICE achieves the highest decoding throughput among
the evaluated methods, with gains across sampling
temperatures, concurrency levels, and hardware
configurations. Our contributions are summarized as follows:
\begin{itemize}
    \item We identify \textbf{revealed sampling randomness}
    as an underexplored source of information for
    allocating speculative computation.

    \item We introduce \textbf{DICE}, a
    \textbf{training-free, lossless} framework that
    exploits this information throughout draft-tree
    construction and in asynchronous prefetching.

    \item We demonstrate \textbf{consistent speedups
    and state-of-the-art decoding throughput}
    across diverse models, tasks, and inference settings.
\end{itemize}

\section{Preliminaries}
\label{sec:prelim}

\paragraph{Draft-Tree Verification.}
Tree-based speculative decoding organizes candidate
continuations into a draft tree for parallel verification
\citep{miao2024specinfer,chen2024sequoia,wang2025opttree}.
Given a prefix $h$, each node $x$ represents a
continuation of length $|x|$, and $xv$ denotes
its child obtained by appending token $v$.
The tree contains at most $N$ non-root nodes.
The target evaluates all candidate prefixes in a single
forward pass.
Verification then follows the target's token choices
from the root until the selected child is absent from
the tree, accepting one token per visited non-root node
and generating one additional token at the stopping node.

We express target sampling using the Gumbel-max
trick~\citep{maddison2014astar}:
\begin{equation}
\textstyle
y_x=\arg\max_v\{z_x(v)+Tg_{|x|+1}(v)\},
\label{eq:target-choice}
\end{equation}
where $z_x$ denotes the target logits after $hx$,
$T>0$ is the sampling temperature, and $g_i$ is
a vector of independent standard Gumbel draws
for the $i$th position after $h$.
This yields the target sampling distribution
$\operatorname{softmax}(z_x/T)$.

\paragraph{Speculative Sampling and Asynchronous Prefetching.}
In standard speculative sampling
\citep{leviathan2023,chen2023}, the drafter proposes
a sequence $x_{1:K}$, which the target evaluates
in a single forward pass and verifies token by token.
A proposal $x_i$ is accepted if
$u_iq_i(x_i)\le p_i(x_i)$, where $p_i$ and $q_i$
are the target and draft distributions at position $i$,
and $u_i\sim\mathrm{Uniform}(0,1)$ is drawn independently
of the proposal.
At the first rejection, verification retains the accepted
prefix and samples a correction token from the residual
distribution proportional to $[p_i-q_i]_+$.
If all proposals are accepted, a bonus token is sampled
from the target distribution.
We denote the verification outcome by $(k,v)$,
where $k$ is the number of accepted draft tokens
and $v$ is the correction or bonus token.
The next round starts from $hx_{1:k}v$.

Speculative speculative decoding (SSD)
\citep{kumar2026ssd} prepares drafts for possible
next-round prefixes while the target verifies
the current proposal.
Each prefetched draft corresponds to a verification
outcome $(k,v)$, and the cache holds at most $B$
entries.
If the actual outcome matches a ready cache entry,
the next round can reuse the prefetched draft
without waiting for another drafting pass.
\section{Method}

In this section, we introduce \textbf{DICE}, a simple
method for further accelerating speculative decoding.
The key technical design is to draw the randomness
used for verification before drafting and synchronize
these draws between the drafting policy and the
target verifier.
The drafting policy combines them with predictions
from an existing draft model to decide which
continuations to prepare, while the verifier uses
the same draws to determine the actual output.
In this way, DICE improves the allocation of
speculative computation without additional
draft-model training or changes to the target
sampling distribution.

We instantiate DICE in two settings:
(i) draft-tree construction (Section~\ref{sec:v3-tree}),
where it guides node selection throughout tree construction,
and (ii) asynchronous prefetching (Section~\ref{sec:v3-prefetch}),
where it prioritizes next-round drafts for prefetching.

\begin{figure}[t]
  \centering
  \includegraphics[width=\textwidth]{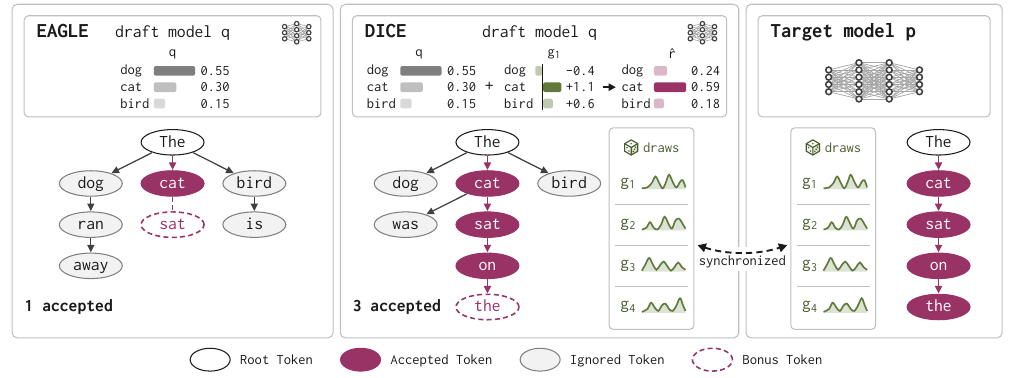}
\caption{\textbf{DICE uses pre-sampled randomness
to guide draft-tree construction.}
Given the same prefix, draft model, and node budget
($N=6$), standard EAGLE constructs its tree using
draft predictions alone, while DICE combines them
with known Gumbel draws to guide node selection
throughout tree construction.
The score insets illustrate how the known
perturbations change token preferences.
The target uses its true logits and the same
synchronized draws to produce the realized
continuation shown on the right.
Colored nodes indicate accepted draft tokens,
while dashed nodes indicate the next target token
absent from the tree.
In this example, DICE accepts three draft tokens
compared with one for EAGLE.}
\label{fig:dice-overview}
\end{figure}

\subsection{Seed-Aware Draft-Tree Construction}
\label{sec:v3-tree}

At each node $x$, the target's next-token choice depends
on its logits $z_x$ and the Gumbel perturbation
$g_{|x|+1}$ (Eq.~\ref{eq:target-choice}).
As discussed earlier, obtaining $z_x$ requires a
target-model forward pass, whereas $g_{|x|+1}$ can
be sampled before drafting.
We combine the draft logits $d_x$ with the known
perturbation to estimate the target's next-token choice:
\begin{equation}
\textstyle
\widehat r_x(v)=
\operatorname{softmax}_v\left(
\frac{\alpha d_x+Tg_{|x|+1}}{\sigma}
\right),
\label{eq:dice-token-score}
\end{equation}
where $\alpha,\sigma>0$ control the relative scale
of draft logits and the concentration of the resulting
scores, respectively.
For tokens with the same draft logit, a larger
Gumbel perturbation yields a higher selection score.
We calibrate these two parameters only once for each
target--drafter pair.
In practice, the first 2,000 nodes observed during decoding
(typically 1--3 prompts) are sufficient for accurate calibration.
The derivation of the scoring rule and an analysis
of calibration accuracy are provided in Appendix~\ref{app:tree-score}.

To score entire continuations, we accumulate the
local selection scores along each path:
\begin{equation}
\textstyle
\widehat w(\varnothing)=1,\qquad
\widehat w(xv)=\widehat w(x)\widehat r_x(v).
\label{eq:dice-path-score}
\end{equation}
Verification reaches a node only if the target selects
every token along its path, so $\widehat w(x)$ estimates
the likelihood that the node will be visited.

We use these scores throughout tree construction.
At each drafting step, the local scores determine
which child tokens to propose, while the cumulative
path scores determine which nodes to expand next.
The same path scores select the final nodes for
target verification.
When dynamic tree sizing is enabled, they also
guide the choice of drafting depth and verification
width.
Algorithm~\ref{alg:dice-tree} summarizes the resulting tree-construction procedure.

This approach is agnostic to the underlying draft
model and applies to tree-based decoding with
standalone small drafters
\citep{svirschevski2024specexec,wang2025opttree},
EAGLE-family drafters
\citep{li2024eagle,li2025eagle3,eagle2026eagle4},
and multi-token prediction (MTP) modules
\citep{gloeckle2024mtp,deepseek2024v3}.
DICE retains their existing drafting architectures
and tree-verification infrastructure.

\begin{algorithm}[t]
\caption{DICE for Draft-Tree Construction}
\label{alg:dice-tree}
\begin{algorithmic}[1]
\REQUIRE Prefix $h$, drafter $D$, policy $\mathcal P$,
budget $N$, steps $L$, calibrated $(\alpha,\sigma)$,
temperature $T$, draws $\{g_i\}$
\ENSURE Verification tree $\mathcal S$, $|\mathcal S|\le N$
\STATE $\mathcal U\gets\mathcal F\gets\{\varnothing\}$,
\quad $\widehat w(\varnothing)\gets1$
\dicestage{Draft and score candidate nodes}
\FOR{$\ell=1,\ldots,L$}
  \STATE $\{d_x\}_{x\in\mathcal F}\gets D(h,\mathcal F)$
  \STATE $\widehat r_x\gets\operatorname{softmax}
  \big((\alpha d_x+Tg_{|x|+1})/\sigma\big)$,
  \quad $x\in\mathcal F$
  \STATE $\mathcal C\gets\operatorname{Propose}_{\mathcal P}
  (\mathcal F,\{\widehat r_x\})$
  \STATE $\widehat w(xv)\gets\widehat w(x)\widehat r_x(v)$,
  \quad $xv\in\mathcal C$
  \STATE $\mathcal U\gets\mathcal U\cup\mathcal C$
  \STATE $\mathcal F\gets\operatorname{Select}_{\mathcal P}
  (\mathcal C,\widehat w)$
\ENDFOR
\dicestage{Select verification nodes}
\STATE $\mathcal S\gets\operatorname{Top}_{N}
(\mathcal U\setminus\{\varnothing\},\widehat w)$
\STATE \textbf{return} $\mathcal S$
\end{algorithmic}
\end{algorithm}

The target then verifies the constructed tree using
its true logits and the same Gumbel draws used
during drafting.
Each Gumbel vector is tied to a logical output
position, so nodes at the same depth share the
corresponding vector.
Unused draws are carried over to subsequent
decoding rounds.
Thus, DICE changes which continuations are computed
in advance, while the generated sequence remains
identical to autoregressive target sampling under
the same random draws.
A formal correctness proof is provided in Appendix~\ref{app:correctness}.

\subsection{Seed-Aware Asynchronous Prefetching}
\label{sec:v3-prefetch}

Beyond draft-tree construction, DICE also guides
which next-round drafts to prefetch.
Each verification outcome $(k,v)$ determines a possible
next-round prefix $hx_{1:k}v$, and the cache can hold
drafts for at most $B$ of them.
DICE therefore estimates which outcomes are most likely
given the pre-sampled verification randomness and
draft-model predictions.
The uniform draws $u_i$ help predict how many draft
tokens will be accepted, while the Gumbel vector
$g_{k+1}$ helps predict the correction or bonus token $v$.

To estimate the accepted length, we model the unknown
target-to-draft probability ratio
$\rho_i=p_i(x_i)/q_i(x_i)$ for each proposed token.
The acceptance condition becomes $\rho_i\ge u_i$.
Since the target probability is unavailable during
prefetching, we estimate the distribution of $\rho_i$
from the draft probability $q_i(x_i)$.
Specifically, we assume that the draft probabilities
are calibrated, so that
$\mathbb E[\rho_i\mid q_i(x_i)]=1$.
Together with the bound
$0\le\rho_i\le 1/q_i(x_i)$, this gives a
maximum-entropy model $f_{q_i(x_i)}$ in the form
of a truncated exponential distribution.
Its parameters are determined by the draft
probability alone, with no additional fitting
(Appendix~\ref{app:maxent}).
Given the known draw $u_i$, we estimate the
acceptance probability as
\begin{equation}
\textstyle
\widehat\eta_i
=\Pr_{\rho\sim f_{q_i(x_i)}}(\rho\ge u_i).
\label{eq:dice-accept-prob}
\end{equation}
Assuming independent ratios across positions,
the probability of accepting exactly $k$ draft
tokens is estimated by
\begin{equation}
\textstyle
\widehat P(k)=
\begin{cases}
\left(\prod_{i=1}^{k}\widehat\eta_i\right)
(1-\widehat\eta_{k+1}), & k<K,\\
\prod_{i=1}^{K}\widehat\eta_i, & k=K.
\end{cases}
\label{eq:dice-stop-prob}
\end{equation}

We next estimate which token follows the accepted
prefix, using the same ratio model and the known
Gumbel vector $g_{k+1}$.
For each possible $k$, we sample ratios
$\rho_{k+1}(v)$ from $f_{q_{k+1}(v)}$ and construct
unnormalized token weights
\begin{equation}
\textstyle
m_k(v)=
\begin{cases}
q_{k+1}(v)[\rho_{k+1}(v)-1]_+, & k<K,\\
q_{K+1}(v)\rho_{K+1}(v), & k=K.
\end{cases}
\label{eq:dice-prefetch-weights}
\end{equation}
The first case corresponds to correction sampling
after rejection, where the rejected proposal has
zero residual weight.
The second corresponds to bonus-token sampling
when all proposals are accepted.
With $g_{k+1}$ held fixed, we repeatedly select
the token maximizing $\log m_k(v)+g_{k+1}(v)$.
The resulting token frequencies estimate
$\widehat r_k(v)$.
Combining both predictions gives the estimated
probability of each verification outcome,
\begin{equation}
\textstyle
\widehat\pi(k,v)=
\widehat P(k)\widehat r_k(v).
\label{eq:dice-prefetch-score}
\end{equation}
DICE selects the $B$ highest-scoring outcomes
and asynchronously prepares their corresponding
drafts using the existing SSD pipeline~\citep{kumar2026ssd}.

Verification uses the true target probabilities
and the same pre-sampled random draws.
For each possible next-round prefix, prefetched
proposals are sampled from the original draft
distribution independently of their own
acceptance uniforms.
A cache hit reuses the prepared draft, while
a miss falls back to regular drafting.
Thus, DICE changes which drafts are prepared
in advance without altering the target
sampling distribution.

\subsection{Implementation}
\label{sec:implementation}

We generate verification randomness on demand using
a deterministic, counter-based hash indexed by the
request seed and absolute token position, with an
additional vocabulary index for Gumbel perturbations.
Separate random streams are used for acceptance,
target sampling, and draft proposals.
This allows drafting and verification workers, even
on different GPUs, to reproduce the same verification
draws without storing or communicating random-number
buffers.
Indexing by absolute position also keeps the draws
consistent across decoding rounds, tree structures,
and execution schedules.
With the draws available locally, we fuse DICE's
scoring into existing GPU
candidate-selection kernels, applying the known
perturbations during vocabulary filtering when present.
The additional operations run within the existing
drafting step and can be captured in CUDA graphs,
without requiring extra model forward passes.

\section{Experiments}
\label{sec:experiments}

This section studies (i) how much DICE improves
decoding throughput and (ii) how consistently
these gains hold across models, tasks, and
inference settings.

\begin{table}[t]
  \centering
  \caption{Tree-based decoding on Spec-Bench and SPEED-Bench at $T=1$
  and concurrency 1 on H200 GPUs.
  Baseline is the drafter's original tree method.
  Multipliers are relative to AR; green percentages give the throughput
  gain over Baseline.}
  \label{tab:tree-results}
  \begingroup
  \fontsize{7.5}{9}\selectfont
  \setlength{\tabcolsep}{1.8pt}
  \renewcommand{\arraystretch}{1.08}
  \setlength{\heavyrulewidth}{0.65pt}
  \setlength{\lightrulewidth}{0.35pt}
  \setlength{\cmidrulewidth}{0.25pt}
  \arrayrulecolor{black!70}
  \sisetup{mode=text, detect-weight=true, detect-family=true,
    table-number-alignment=center}
  \definecolor{dicegain}{RGB}{0,115,65}
  \newcommand{\ssgain}[1]{\textcolor{dicegain}{\fontsize{6.5}{8}\selectfont(+#1\%)}}
  \newlength{\sstargetwidth}
  \newlength{\ssdraftwidth}
  \newlength{\ssbackendwidth}
  \newlength{\ssratewidth}
  \newlength{\ssfactorwidth}
  \newlength{\sstauwidth}
  \newlength{\ssgainwidth}
  \settowidth{\sstargetwidth}{Llama-3.1-70B}
  \settowidth{\ssdraftwidth}{Nemo3.5-120B}
  \ifdim\ssdraftwidth>\sstargetwidth\setlength{\sstargetwidth}{\ssdraftwidth}\fi
  \settowidth{\ssdraftwidth}{EAGLE-4}
  \settowidth{\ssbackendwidth}{FastMTP}
  \ifdim\ssbackendwidth>\ssdraftwidth\setlength{\ssdraftwidth}{\ssbackendwidth}\fi
  \settowidth{\ssbackendwidth}{Llama-3.2-1B}
  \ifdim\ssbackendwidth>\ssdraftwidth\setlength{\ssdraftwidth}{\ssbackendwidth}\fi
  \settowidth{\ssbackendwidth}{Backend}
  \settowidth{\ssratewidth}{SGLang}
  \ifdim\ssratewidth>\ssbackendwidth\setlength{\ssbackendwidth}{\ssratewidth}\fi
  \settowidth{\ssratewidth}{000.0}
  \settowidth{\ssfactorwidth}{$\times$AR}
  \settowidth{\sstauwidth}{0.00}
  \settowidth{\ssgainwidth}{\ssgain{00.00}}
  \newsavebox{\ssmaintable}

  \begin{lrbox}{\ssmaintable}
  \begin{tabular}{@{}
    >{\raggedright\arraybackslash}p{\sstargetwidth}
    @{\hspace{1.5pt}}c@{\hspace{1.5pt}}
    >{\raggedright\arraybackslash}p{\ssdraftwidth}
    >{\centering\arraybackslash}p{\ssbackendwidth}@{\hspace{5pt}}
    S[table-format=3.1,table-column-width=\ssratewidth]@{\hspace{5pt}}
    S[table-format=3.1,table-column-width=\ssratewidth]
    S[table-format=1.2,table-column-width=\ssfactorwidth]
    S[table-format=1.2,table-column-width=\sstauwidth]@{\hspace{5pt}}
    S[table-format=3.1,table-column-width=\ssratewidth]
    S[table-format=1.2,table-column-width=\ssfactorwidth]
    S[table-format=1.2,table-column-width=\sstauwidth]@{\hspace{5pt}}
    >{\raggedleft\arraybackslash}p{\ssgainwidth}@{\hspace{9pt}}
    S[table-format=3.1,table-column-width=\ssratewidth]@{\hspace{5pt}}
    S[table-format=3.1,table-column-width=\ssratewidth]
    S[table-format=1.2,table-column-width=\ssfactorwidth]
    S[table-format=1.2,table-column-width=\sstauwidth]@{\hspace{5pt}}
    S[table-format=3.1,table-column-width=\ssratewidth]
    S[table-format=1.2,table-column-width=\ssfactorwidth]
    S[table-format=1.2,table-column-width=\sstauwidth]@{\hspace{5pt}}
    >{\raggedleft\arraybackslash}p{\ssgainwidth}@{}}
    \toprule
    \multicolumn{4}{@{}l}{}
      & \multicolumn{8}{c}{\textbf{Spec-Bench}}
      & \multicolumn{8}{c@{}}{\textbf{SPEED-Bench}} \\
    \cmidrule(lr){5-12}\cmidrule(l){13-20}
    \multirow{2}{*}[-2.1pt]{Target} & \multirow{2}{*}[-2.1pt]{/}
      & \multirow{2}{*}[-2.1pt]{Drafter}
      & \multirow{2}{*}[-2.1pt]{Backend} & {AR}
      & \multicolumn{3}{@{}c@{\hspace{5pt}}}{Baseline}
      & \multicolumn{3}{@{}c@{\hspace{5pt}}}{DICE} & \multicolumn{1}{@{}c@{\hspace{9pt}}}{Gain}
      & {AR} & \multicolumn{3}{@{}c@{\hspace{5pt}}}{Baseline}
      & \multicolumn{3}{@{}c@{\hspace{5pt}}}{DICE} & \multicolumn{1}{@{}c@{}}{Gain} \\
    \cmidrule(l{1pt}r{1pt}){5-5}\cmidrule(l{1pt}r{1pt}){6-8}\cmidrule(l{1pt}r{1pt}){9-11}
    \cmidrule(l{1pt}r{1pt}){13-13}\cmidrule(l{1pt}r{1pt}){14-16}\cmidrule(l{1pt}r{1pt}){17-19}
      & & & & {tok/s} & {tok/s} & {$\times$AR} & {$\tau$}
      & {tok/s} & {$\times$AR} & {$\tau$} & \multicolumn{1}{@{}c@{\hspace{9pt}}}{tok/s}
      & {tok/s} & {tok/s} & {$\times$AR} & {$\tau$}
      & {tok/s} & {$\times$AR} & {$\tau$} & \multicolumn{1}{@{}c@{}}{tok/s} \\
    \midrule
    Vicuna-33B & / & Vicuna-7B & HF
      & 24.2 & 25.8 & 1.07 & 5.11 & 28.3 & 1.17 & 5.54 & \ssgain{9.64}
      & 24.4 & 26.0 & 1.06 & 5.09 & 28.9 & 1.18 & 5.57 & \ssgain{11.46} \\
    Qwen2-72B & / & Qwen2-7B & HF
      & 13.0 & 20.3 & 1.55 & 5.57 & 21.1 & 1.62 & 6.16 & \ssgain{4.34}
      & 13.4 & 19.3 & 1.44 & 4.64 & 20.8 & 1.55 & 5.44 & \ssgain{8.06} \\
    Llama-3.1-70B & / & Llama-3.2-1B & HF
      & 13.0 & 26.9 & 2.08 & 5.41 & 28.1 & 2.17 & 6.78 & \ssgain{4.34}
      & 12.7 & 23.1 & 1.81 & 4.13 & 26.8 & 2.10 & 5.78 & \ssgain{15.95} \\
    Llama-3.1-70B & / & Llama-3.1-8B & HF
      & 12.5 & 20.3 & 1.63 & 7.18 & 21.2 & 1.70 & 8.59 & \ssgain{4.50}
      & 13.0 & 17.2 & 1.32 & 4.37 & 21.2 & 1.63 & 7.31 & \ssgain{23.55} \\
    DeepSeek-70B & / & DeepSeek-8B & HF
      & 12.7 & 18.1 & 1.43 & 6.56 & 19.8 & 1.56 & 6.81 & \ssgain{9.45}
      & 12.5 & 17.6 & 1.40 & 5.97 & 19.5 & 1.56 & 6.25 & \ssgain{11.14} \\
    \arrayrulecolor{black!35}\cmidrule[0.25pt]{1-20}\arrayrulecolor{black!70}
    Qwen2-7B & / & EAGLE-2 & HF
      & 53.7 & 99.8 & 1.86 & 3.05
      & 107.1 & 2.00 & 3.39 & \ssgain{7.20}
      & 53.2 & 89.0 & 1.67 & 2.73
      & 100.3 & 1.89 & 3.20 & \ssgain{12.70} \\
    Qwen2-72B & / & EAGLE-2 & HF
      & 9.1 & 27.5 & 3.01 & 3.52
      & 29.3 & 3.21 & 3.80 & \ssgain{6.70}
      & 8.6 & 23.5 & 2.74 & 3.14
      & 25.1 & 2.93 & 3.42 & \ssgain{6.76} \\
    Mixtral-8$\times$7B & / & EAGLE-2 & HF
      & 17.1 & 26.2 & 1.54 & 3.25
      & 27.8 & 1.63 & 3.46 & \ssgain{5.90}
      & 16.1 & 25.1 & 1.56 & 3.27
      & 26.6 & 1.65 & 3.49 & \ssgain{5.64} \\
    \arrayrulecolor{black!35}\cmidrule[0.25pt]{1-20}\arrayrulecolor{black!70}
    Llama-3.1-8B & / & EAGLE-3 & HF
      & 34.7 & 81.1 & 2.33 & 3.78
      & 98.6 & 2.84 & 4.74 & \ssgain{21.50}
      & 32.4 & 53.1 & 1.64 & 2.83
      & 66.2 & 2.05 & 3.62 & \ssgain{24.62} \\
    Llama-3.1-70B & / & EAGLE-3 & HF
      & 10.8 & 35.4 & 3.28 & 4.07
      & 41.6 & 3.85 & 4.86 & \ssgain{17.80}
      & 10.7 & 24.8 & 2.32 & 2.82
      & 29.3 & 2.74 & 3.38 & \ssgain{18.00} \\
    DeepSeek-8B & / & EAGLE-3 & HF
      & 32.3 & 88.8 & 2.75 & 4.40
      & 96.5 & 2.98 & 4.89 & \ssgain{8.70}
      & 32.5 & 78.6 & 2.42 & 3.86
      & 84.4 & 2.60 & 4.25 & \ssgain{7.43} \\
    \arrayrulecolor{black!35}\cmidrule[0.25pt]{1-20}\arrayrulecolor{black!70}
    Qwen3.8-27B & / & MTP & SGLang
      & 136.3 & 299.8 & 2.20 & 3.39 & 320.3 & 2.35 & 4.13 & \ssgain{6.87}
      & 131.0 & 284.8 & 2.17 & 3.21 & 311.0 & 2.37 & 3.94 & \ssgain{9.22} \\
    Qwen3.6-35B & / & MTP & SGLang
      & 222.7 & 466.1 & 2.09 & 3.65 & 516.7 & 2.32 & 4.48 & \ssgain{10.86}
      & 240.8 & 479.0 & 1.99 & 3.46 & 527.9 & 2.19 & 4.28 & \ssgain{10.20} \\
    MiMo-7B & / & MTP & SGLang
      & 171.7 & 363.5 & 2.12 & 3.79 & 381.6 & 2.22 & 3.91 & \ssgain{4.97}
      & 170.6 & 340.5 & 2.00 & 3.27 & 361.5 & 2.12 & 3.71 & \ssgain{6.15} \\
    \arrayrulecolor{black!35}\cmidrule[0.25pt]{1-20}\arrayrulecolor{black!70}
    Qwen3.8-27B & / & EAGLE-4 & SGLang
      & 127.1 & 405.2 & 3.19 & 4.29 & 451.9 & 3.55 & 4.63 & \ssgain{11.52}
      & 128.7 & 370.4 & 2.88 & 3.71 & 406.9 & 3.16 & 4.08 & \ssgain{9.86} \\
    Qwen3.8-27B & / & EAGLE-4 & vLLM
      & 166.8 & 428.1 & 2.57 & 3.76 & 491.9 & 2.95 & 4.34 & \ssgain{14.91}
      & 169.6 & 377.6 & 2.23 & 3.28 & 456.4 & 2.69 & 3.94 & \ssgain{20.88} \\
    Nemo3.5-120B & / & EAGLE-4 & SGLang
      & 129.2 & 362.7 & 2.81 & 5.53 & 404.4 & 3.13 & 5.80 & \ssgain{11.49}
      & 132.1 & 351.0 & 2.66 & 4.85 & 383.0 & 2.90 & 5.00 & \ssgain{9.14} \\
    Nemo3.5-120B & / & EAGLE-4 & vLLM
      & 207.8 & 444.2 & 2.14 & 5.55 & 485.2 & 2.34 & 5.96 & \ssgain{9.21}
      & 210.5 & 405.2 & 1.93 & 4.81 & 448.0 & 2.13 & 5.19 & \ssgain{10.56} \\
    \bottomrule
  \end{tabular}
  \end{lrbox}
  \ifdim\wd\ssmaintable>\linewidth\sbox{\ssmaintable}{\resizebox{\linewidth}{!}{\usebox{\ssmaintable}}}\fi
  \usebox{\ssmaintable}

  \par\vspace{3pt}
  \begin{minipage}{\wd\ssmaintable}
    \fontsize{7}{8.2}\selectfont
    HF: Hugging Face Transformers.
    DeepSeek-70B and DeepSeek-8B denote DeepSeek-R1-Distill-Llama-70B and -8B,
    and Qwen3.6-35B and Nemo3.5-120B denote Qwen3.6-35B-A3B and
    Nemotron-3.5-120B-A12B.
  \end{minipage}
  \endgroup
\end{table}

\begin{table}[t]
  \centering
  \caption{Asynchronous prefetching on Alpaca, GSM8K, HumanEval, and
  UltraFeedback at $T=1$ and concurrency 1 on H200 GPUs.
  Baseline is vanilla SSD.
  Multipliers are relative to AR; Gain reports the throughput
  improvement over Baseline.}
  \label{tab:ssd-results}
  \begingroup
  \fontsize{8}{9.6}\selectfont
  \setlength{\tabcolsep}{1.8pt}
  \renewcommand{\arraystretch}{1.03}
  \sisetup{mode=text, detect-weight=true, detect-family=true,
    table-number-alignment=center}
  \definecolor{dicegain}{RGB}{0,115,65}
  \newcommand{\ssdgain}[1]{\textcolor{dicegain}{\fontsize{7}{8.4}\selectfont(+#1\%)}}
  \newlength{\ssdratewidth}
  \newlength{\ssdfactorwidth}
  \newlength{\ssdtauwidth}
  \newlength{\ssdhitwidth}
  \newlength{\ssdgainwidth}
  \settowidth{\ssdratewidth}{000.0}
  \settowidth{\ssdfactorwidth}{$\times$AR}
  \settowidth{\ssdtauwidth}{0.00}
  \settowidth{\ssdhitwidth}{00.0}
  \settowidth{\ssdgainwidth}{\ssdgain{00.00}}
  \newsavebox{\ssdasynctable}

  \begin{lrbox}{\ssdasynctable}
  \begin{tabular}{@{}
    l@{\hspace{2pt}}c@{\hspace{2pt}}l@{\hspace{8pt}}
    S[table-format=3.1,table-column-width=\ssdratewidth]@{\hspace{8pt}}
    S[table-format=3.1,table-column-width=\ssdratewidth]
    S[table-format=1.2,table-column-width=\ssdfactorwidth]
    S[table-format=1.2,table-column-width=\ssdtauwidth]
    S[table-format=2.1,table-column-width=\ssdhitwidth]@{\hspace{8pt}}
    S[table-format=3.1,table-column-width=\ssdratewidth]
    S[table-format=1.2,table-column-width=\ssdfactorwidth]
    S[table-format=1.2,table-column-width=\ssdtauwidth]
    S[table-format=2.1,table-column-width=\ssdhitwidth]@{\hspace{8pt}}
    >{\raggedleft\arraybackslash}p{\ssdgainwidth}@{}}
    \toprule
    \multirow{2}{*}[-2.1pt]{Target} & \multirow{2}{*}[-2.1pt]{/}
      & \multirow{2}{*}[-2.1pt]{Drafter}
      & {AR} & \multicolumn{4}{@{}c@{\hspace{8pt}}}{Baseline}
      & \multicolumn{4}{@{}c@{\hspace{8pt}}}{DICE}
      & \multicolumn{1}{@{}c@{}}{Gain} \\
    \cmidrule(l{1pt}r{1pt}){4-4}\cmidrule(l{1pt}r{1pt}){5-8}
    \cmidrule(l{1pt}r{1pt}){9-12}\cmidrule(l{1pt}r{0pt}){13-13}
      & & & {tok/s} & {tok/s} & {$\times$AR} & {$\tau$} & {Hit}
      & {tok/s} & {$\times$AR} & {$\tau$} & {Hit}
      & \multicolumn{1}{@{}c@{}}{tok/s} \\
    \midrule
    Llama-3.1-70B-Instruct & / & Llama-3.2-1B-Instruct
      & 66.9 & 160.8 & 2.40 & 3.85 & 59.8
      & 182.0 & 2.72 & 3.97 & 77.9 & \ssdgain{13.21} \\
    Qwen3-32B & / & Qwen3-0.6B
      & 110.2 & 165.3 & 1.50 & 2.94 & 67.6
      & 172.0 & 1.56 & 2.90 & 75.2 & \ssdgain{4.03} \\
    Llama-3.1-70B (Base) & / & Llama-3.2-1B (Base)
      & 68.4 & 123.7 & 1.81 & 3.19 & 45.6
      & 140.6 & 2.06 & 3.21 & 69.8 & \ssdgain{13.65} \\
    Llama-3.1-8B-Instruct & / & Llama-3.2-1B-Instruct
      & 187.1 & 238.1 & 1.27 & 2.86 & 56.9
      & 266.8 & 1.43 & 2.87 & 80.7 & \ssdgain{12.04} \\
    \bottomrule
  \end{tabular}
  \end{lrbox}
  \ifdim\wd\ssdasynctable>\linewidth\sbox{\ssdasynctable}{\resizebox{\linewidth}{!}{\usebox{\ssdasynctable}}}\fi
  \usebox{\ssdasynctable}

  \par\vspace{3pt}
  \begin{minipage}{\wd\ssdasynctable}
    \fontsize{7.5}{8.8}\selectfont
    Hit is the outcome-cache hit rate (\%).
  \end{minipage}
  \endgroup
\end{table}

\begin{figure}[t]
  \centering
  \includegraphics[width=\textwidth]{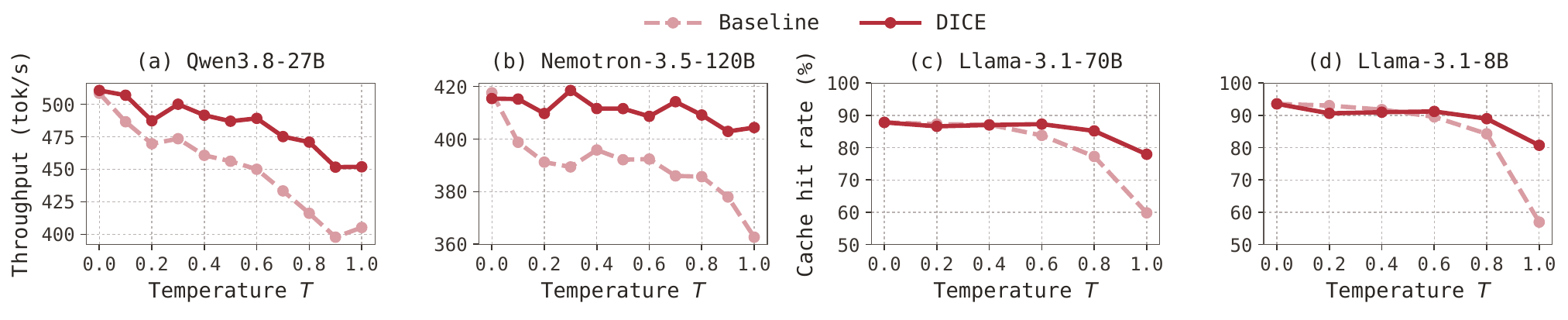}
  \caption{Throughput across sampling temperatures.
  (a, b) Draft trees in SGLang on Spec-Bench, where Baseline is EAGLE-4.
  (c, d) Cache hit rate of asynchronous prefetching with a
  Llama-3.2-1B-Instruct drafter, where Baseline is SSD.}
  \label{fig:temperature}
\end{figure}

\begin{figure}[t]
  \centering
  \includegraphics[width=\textwidth]{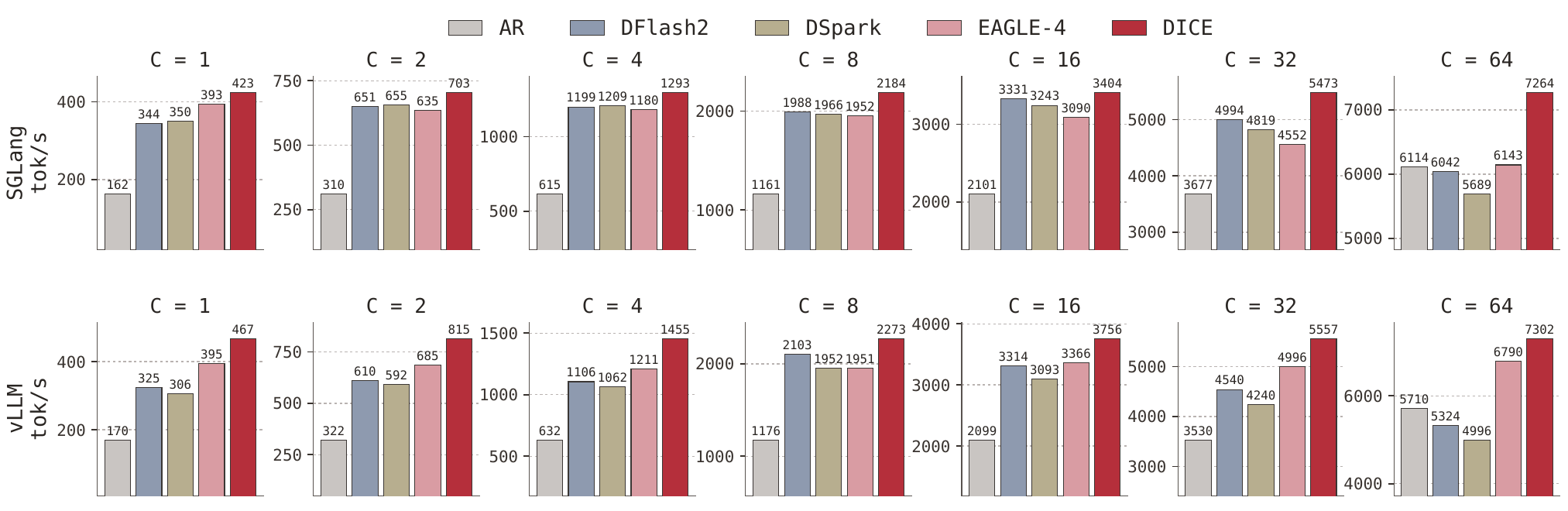}
  \caption{Throughput across concurrency levels $C$ for Qwen3.8-27B on
  SPEED-Bench, in SGLang (top) and vLLM (bottom).}
  \label{fig:concurrency}
\end{figure}

\subsection{Experimental Setup}

\paragraph{Models and Benchmarks.}
For draft-tree decoding, we evaluate DICE on
16 target--drafter pairs, including five with
standalone small drafters using OPT-Tree
\citep{wang2025opttree}, eight with EAGLE-2
\citep{li2024eagle2}, EAGLE-3
\citep{li2025eagle3}, or EAGLE-4
\citep{eagle2026eagle4}, and three with
multi-token prediction (MTP) modules
\citep{gloeckle2024mtp,deepseek2024v3}.
The two EAGLE-4 pairs are evaluated in both
SGLang~\citep{zheng2024sglang} and
vLLM~\citep{kwon2023vllm}, yielding 18 model--backend
configurations.
For asynchronous prefetching, we evaluate DICE
with SSD~\citep{kumar2026ssd} on four
target--drafter pairs, including two from the
original study and two additional settings.

The draft-tree experiments use
Spec-Bench~\citep{xia2024specbench} and
SPEED-Bench~\citep{abramovich2026speedbench},
covering conversation, translation, summarization,
mathematical reasoning, question answering,
and retrieval-augmented generation.
The prefetching experiments follow the original
SSD evaluation suite of
Alpaca~\citep{taori2023alpaca},
GSM8K~\citep{cobbe2021gsm8k},
HumanEval~\citep{chen2021humaneval}, and
UltraFeedback~\citep{cui2024ultrafeedback}.

\paragraph{Baselines and Backends.}
We compare DICE against the corresponding original
speculative decoding methods, keeping the draft-model
weights and inference backend unchanged.
For draft-tree decoding, we use reference implementations
based on Hugging Face Transformers~\citep{wolf2020transformers}
for standalone drafters and EAGLE-2/3, and optimized
implementations in SGLang~\citep{zheng2024sglang}
and vLLM~\citep{kwon2023vllm} for MTP and EAGLE-4.
For asynchronous prefetching, we use the original
SSD implementation~\citep{kumar2026ssd},
replacing its prefetch-selection policy with DICE.
We also include autoregressive decoding as a reference
for measuring speedup.

\paragraph{Evaluation.}
We report decoding throughput, its speedup over autoregressive
decoding (AR), and the average acceptance length $\tau$, defined as
the number of tokens generated per target forward pass.
For prefetching, we also report the cache hit rate.
Each baseline uses the configuration recommended by its authors,
or the fastest configuration found on a held-out split when no
recommendation is given or a faster one exists.
DICE is configured in the same way.
Unless otherwise stated, we sample at $T=1$ and serve one request
at a time on H200 GPUs.

\subsection{Draft-Tree Decoding}
\label{sec:tree-results}

\paragraph{Main results.}
Table~\ref{tab:tree-results} shows that DICE improves throughput for
every target--drafter pair and backend, by 4.3--21.5\% on Spec-Bench
and 5.6--24.6\% on SPEED-Bench.
The gains hold for standalone drafters, EAGLE drafters, and MTP
modules, and in both reference and production backends.
In every case, DICE accepts more tokens per target forward pass,
since the known randomness steers the tree toward the continuation
the target will sample.

\paragraph{Sampling temperature.}
Figure~\ref{fig:temperature}(a, b) compares EAGLE-4 and DICE across
sampling temperatures at a fixed node budget.
At $T=0$, both methods verify greedily and run at the same speed.
As the temperature rises, the target distribution becomes more
diffuse, so a fixed node budget covers the sampled continuation less
often, and the throughput of EAGLE-4 drops.
DICE builds the tree for the draws that will actually be used,
so its throughput declines much less and its advantage grows with the
temperature.

\begin{figure}[t]
  \centering
  \includegraphics[width=\textwidth]{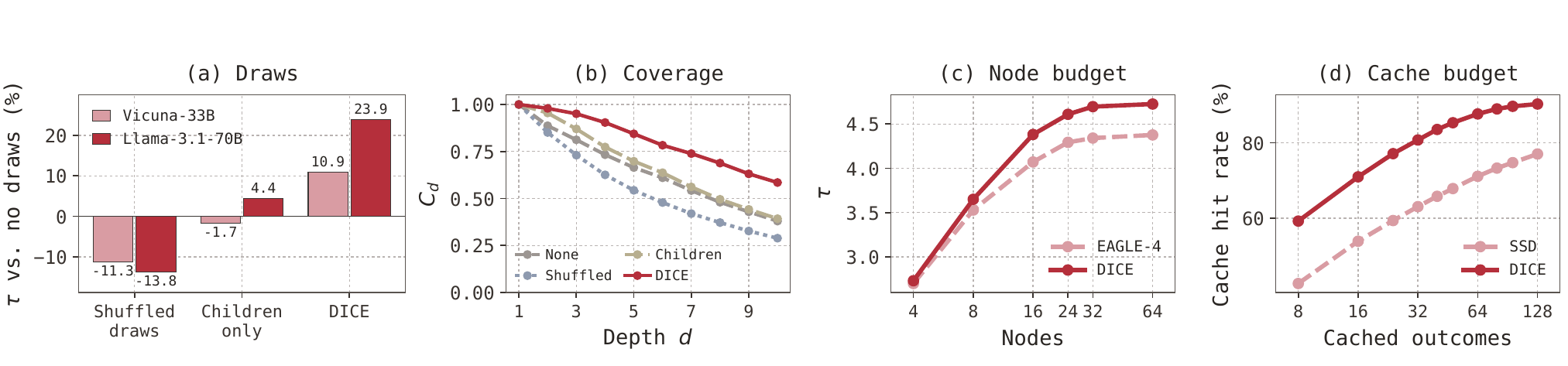}
  \caption{Where the gains come from.
  (a) Change in acceptance length over a drafter without draws, on
  fixed trees of 60 nodes and depth 10 (OPT-Tree, Spec-Bench).
  Shuffled draws are independent of the target's draws, and children
  only uses the draws to pick each node's children but allocates the
  budget by the draft probabilities.
  (b) Probability $C_d$ that the tree covers the target's continuation
  through depth $d$ (Llama-3.1-70B with a Llama-3.1-8B drafter).
  (c) Acceptance length versus node budget (Qwen3.8-27B with EAGLE-4 in
  SGLang). (d) Cache hit rate versus cache budget
  (Llama-3.1-70B-Instruct with SSD).}
  \label{fig:sources}
\end{figure}

\paragraph{Concurrency.}
Serving many requests at once leaves less spare compute for
speculation.
Figure~\ref{fig:concurrency} reports the throughput of DICE, EAGLE-4,
and two parallel drafters, DFlash2~\citep{chen2026dflash} and
DSpark~\citep{cheng2026dspark}, as concurrency grows from 1 to 64.
DICE is the fastest method at every concurrency in both SGLang and
vLLM, ahead of the strongest baseline by 2--18\% and 8--20\%,
respectively.
At concurrency 64, DFlash2 and DSpark fall below the speed of AR,
whereas DICE remains 1.19$\times$ and 1.28$\times$ faster than AR.
When the engine shortens speculation to a single draft chain at high
concurrency, DICE applies unchanged, since a chain is a tree with one
child per node.

\subsection{Asynchronous Prefetching}
\label{sec:prefetch-results}

\wrapneedspace{13}
\begin{wrapfigure}{r}{0.5\linewidth}
  \vspace{-10pt}
  \centering
  \includegraphics[width=\linewidth]{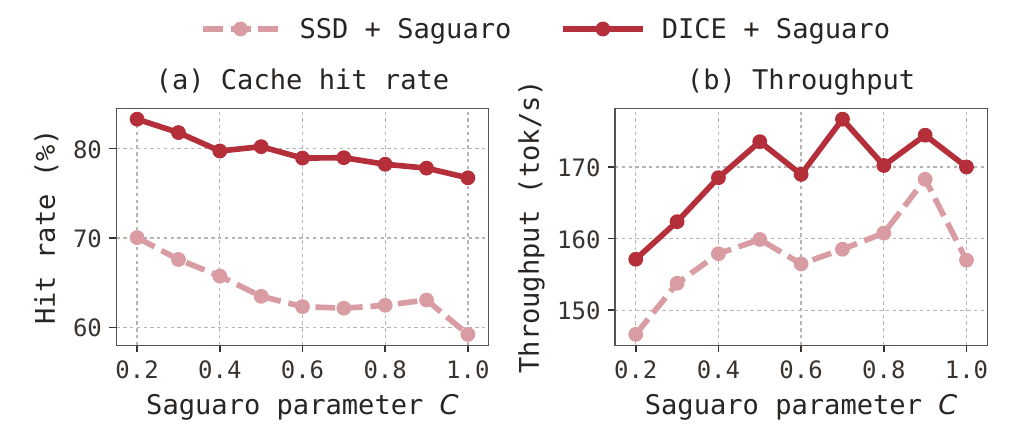}
  \caption{SSD with Saguaro sampling, with and without DICE, for
  Llama-3.1-70B-Instruct with a Llama-3.2-1B-Instruct drafter.
  Smaller $C$ reshapes the drafts more strongly, and $C=1$ is
  standard sampling.}
  \label{fig:saguaro}
  \vspace{-6pt}
\end{wrapfigure}
\paragraph{Main results.}
Table~\ref{tab:ssd-results} shows that DICE raises the cache hit rate
of SSD on every pair, from 45.6--67.6\% to 69.8--80.7\%, which yields
4.0--13.7\% higher throughput.
The acceptance length stays nearly unchanged, because DICE changes
which drafts are prepared but not the drafts themselves.
The gains are largest where vanilla SSD misses most often.
Figure~\ref{fig:temperature}(c, d) shows how this changes with the
temperature.
At low temperatures, SSD already predicts most verification outcomes
and DICE matches it.
As sampling becomes more random, the hit rate of SSD falls to 57--60\%
at $T=1$, whereas DICE keeps it near 80\%.

\paragraph{Comparison with Saguaro sampling.}
SSD also raises its hit rate with Saguaro sampling~\citep{kumar2026ssd},
which reshapes the draft distribution with a parameter $C$ so that the
next verification outcome becomes easier to predict, at the cost of
lower acceptance.
DICE raises the hit rate without changing the drafts, so the two
approaches are complementary.
Figure~\ref{fig:saguaro} sweeps $C$.
At every $C$, DICE adds 13--18 points of hit rate.
DICE alone ($C=1$) matches SSD at its best $C$
(170 versus 168 tokens/s), and combining the two reaches
177 tokens/s.

\section{Discussion}
\label{sec:discussion}

As shown in Section~\ref{sec:experiments}, DICE achieves consistent
speedups across drafters, inference engines, and sampling settings
with only a simple change to how speculative computation is allocated.
To better understand why, this section examines (i) where these gains
come from and (ii) how much they depend on modeling the target's
unknown predictions.

\subsection{Where Do the Gains Come From?}

Could the gains come simply from perturbing the draft scores?
DICE adds Gumbel noise to the drafter's logits, and noise alone already
changes which candidates a tree explores.
To separate the information in the draws from the perturbation, we fix
the tree budget and the target's draws and change only what the
drafter sees (Figure~\ref{fig:sources}a).
With shuffled draws, the drafter follows the scoring rule of DICE but
with draws that are independent of the ones the target will use.
This lowers the acceptance length by 11--14\% relative to a drafter
that sees no draws.
With synchronized draws, DICE raises it by 11--24\%.
The gain comes from knowing the draws that the target will actually
use, not from randomizing the drafter.

This knowledge helps only when it shapes the allocation.
If DICE chooses each node's children but the tree still spends its
budget according to the draft probabilities, most of the gain disappears
(Figure~\ref{fig:sources}a).
Figure~\ref{fig:sources}b shows where the budget goes.
It plots the probability $C_d$ that the tree contains the target's
continuation through depth $d$, whose sum over $d$ is the acceptance
length.
DICE covers the realized path more often at every depth beyond the
first, and the gap widens with depth.

A better allocation should matter more when there is more to allocate.
Figure~\ref{fig:sources}c and d vary the budget.
For draft trees, the advantage of DICE grows with the number of nodes,
and EAGLE-4 needs 64 nodes to reach the acceptance length that DICE
attains with 16.
For prefetching, DICE with 8 cached outcomes matches the hit rate that
SSD reaches with 24.
The two kinds of draws contribute differently here.
The acceptance draws $u_i$ indicate where verification will stop, and
the Gumbel draws $g_{k+1}$ indicate which token will follow.
On Llama-3.1-70B-Instruct, either kind alone raises the hit rate of SSD
from 59.8\% to 66.9--67.9\%, and together they reach 76.2\%.

\subsection{How Much Modeling Is Needed?}

\wrapneedspace{13}
\begin{wrapfigure}{r}{0.5\linewidth}
  \vspace{-10pt}
  \centering
  \includegraphics[width=\linewidth]{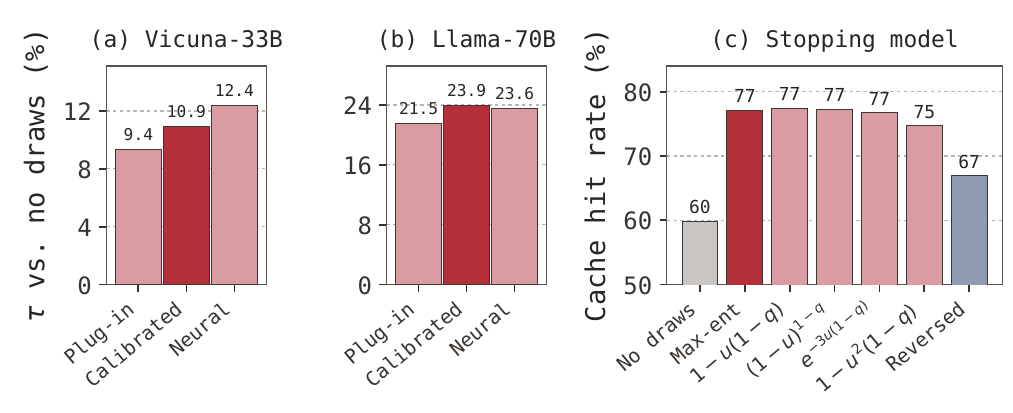}
  \caption{How much modeling is needed.
  (a, b) Change in acceptance length over a drafter without draws for
  three predictors of the target's choices, on the fixed trees of
  Figure~\ref{fig:sources}a (Vicuna-33B and Llama-3.1-70B).
  (c) Cache hit rate of prefetching with different stopping models
  (Llama-3.1-70B-Instruct, 24 cached outcomes).}
  \label{fig:modeling}
  \vspace{-6pt}
\end{wrapfigure}
The preceding results show that the actual draws are informative for
allocating speculative computation.
DICE still relies on estimates of the target's unknown predictions,
however.
How accurate must these estimates be?

For draft trees, we compare predictors of increasing capacity on the
same fixed trees (Figure~\ref{fig:modeling}a, b).
The plug-in rule with $\alpha=\sigma=1$ already raises the acceptance
length by 9--22\%.
Calibrating the two parameters adds one to two points, and a few
hundred observed nodes suffice (Appendix~\ref{app:tree-score}).
A neural predictor that adds a learned correction to the calibrated
score barely changes the held-out likelihood of the target's choices
(0.94 versus 0.95 nats on Vicuna-33B and 0.56 versus 0.55 on
Llama-3.1-70B), and adds at most 1.5 points.
DICE needs estimates that rank candidates well enough to allocate the
budget, not a complete model of the target distribution.

For prefetching, we replace the maximum-entropy model of the stopping
position with other estimates that increase with the observed draws
(Figure~\ref{fig:modeling}c).
These variants change only how the stopping position is estimated, and
all of them reach a hit rate within three points of maximum entropy.
Ignoring the draws or reversing their direction instead loses 10 to 17
points.

Across both applications, the gains depend more on using the actual
verification draws than on the precise form of the probability
estimates.
Even when the target's predictions are known only approximately,
revealing the sampling randomness lets speculative decoding spend its
computation where it will be used.

\section{Related Work}

\paragraph{Speculative decoding.}
Speculative decoding accelerates autoregressive generation by
verifying multiple draft tokens in a single target forward pass
while preserving the target distribution
\citep{leviathan2023,chen2023}.
Subsequent work improves both draft prediction and candidate
organization.
SpecInfer verifies tree-structured candidates
\citep{miao2024specinfer}, while EAGLE-2 uses draft confidence
to construct context-dependent trees \citep{li2024eagle2}.
OPT-Tree explicitly optimizes expected acceptance length
\citep{wang2025opttree}, and Sequoia considers tree structure
together with hardware-dependent costs \citep{chen2024sequoia}.
EAGLE-3 improves draft prediction through multi-layer features
and training-time rollout \citep{li2025eagle3}.
More recently, EAGLE-4 combines an efficient drafter with
cost-aware tree sizing \citep{eagle2026eagle4}.
DICE complements these developments by conditioning candidate
scores on the randomness used in the upcoming verification.

\paragraph{Asynchronous decoding.}
Asynchronous methods overlap drafting with target verification
to reduce sequential execution.
SwiftSpec develops asynchronous drafting together with
KV-cache management and execution optimizations
\citep{zhang2025swiftspec}.
Speculative Speculative Decoding (SSD) prepares future drafts
for predicted verification outcomes \citep{kumar2026ssd}.
DICE builds on this setting by using the actual acceptance
uniforms to predict rejection positions and the subsequent
sampling noise to rank correction or bonus tokens.

\paragraph{Shared randomness.}
Shared-randomness coupling provides an alternative to
rejection-based speculative sampling.
\citet{daliri2025coupling} use communication-free coupling
to obtain drafter-invariant speculative decoding, with outputs
fixed by the target model and the random seed.
\citet{rowan2025listlevel} extend coupling to multiple drafts
and establish conditional drafter invariance.
\citet{kobus2025exponential} introduce exponential-race
speculative decoding and analyze tree construction using
marginal node acceptance probabilities.
PTP introduces sampling randomness
as auxiliary model inputs and trains a model to jointly predict multiple future tokens \citep{draxler2026ptp}.
In contrast, DICE uses shared randomness not only to couple
samples but also to guide speculative computation before
verification outcomes are known. 
This unlocks the broader potential of seed awareness, enabling
general training-free acceleration with straightforward integration into
modern LLM serving engines.

\section{Conclusion}

We presented DICE, a training-free method that removes one source of
uncertainty from speculative decoding.
The random draws that govern verification can be sampled before
drafting and shared between the drafter and the verifier.
DICE combines these known draws with predictions from an existing
drafter to decide which continuations to prepare, guiding node
selection in draft-tree construction and ranking next-round drafts in
asynchronous prefetching.
The target sampling distribution is left unchanged.
Across standalone, EAGLE, and MTP drafters, reference and production
engines, and a range of temperatures and concurrency levels, DICE
consistently improves decoding throughput, by up to 24.6\% in
draft-tree decoding and 13.7\% in asynchronous prefetching.
These gains come from spending the same speculation budget on the
candidates that the sampling run will actually use.
More broadly, our results suggest that sampling randomness need not be
treated as unknown when deciding what to compute in advance, which may
benefit other forms of speculative and asynchronous inference.

\section*{Acknowledgements}
This research was supported in part by Impossible Inc., Jane Street, and Amazon AGI.

\clearpage
\bibliography{references}

\clearpage
\appendix
\etocdepthtag.toc{appendix}
\phantomsection
\section*{Appendix Contents}
\begingroup
\small
\etocsettagdepth{maintext}{none}
\etocsettagdepth{appendix}{subsubsection}
\etocsettocstyle{}{}
\tableofcontents
\endgroup
\newpage

\section{Derivation of the Tree Scoring Rule}
\label{app:tree-score}

\paragraph{A random-utility model of the target choice.}
At node $x$, the target selects
$y_x=\arg\max_v\{z_x(v)+Tg_{|x|+1}(v)\}$
(Eq.~\ref{eq:target-choice}).
During drafting, the draws $g_{|x|+1}$ are known but the target
logits $z_x$ are not.
We model the target logits through the draft logits $d_x$ as
\begin{equation}
z_x(v)=\alpha\, d_x(v)+\epsilon_x(v)+c_x,
\label{eq:app-rum}
\end{equation}
where $c_x$ is a constant shared by all tokens and the errors
$\epsilon_x(v)$ are independent Gumbel variables with scale $\sigma$,
independent of the draws.
The target choice then maximizes a known utility plus independent
Gumbel noise,
\begin{equation}
y_x=\arg\max_v\{\alpha\, d_x(v)+Tg_{|x|+1}(v)+\epsilon_x(v)\}.
\end{equation}
By the Gumbel-max property, which is also the basis of the
multinomial logit model~\citep{mcfadden1974conditional},
\begin{equation}
\Pr\bigl(y_x=v\mid d_x,g_{|x|+1}\bigr)
=\operatorname{softmax}_v\!\left(\frac{\alpha d_x+Tg_{|x|+1}}{\sigma}\right)
=\widehat r_x(v),
\end{equation}
which is Eq.~\eqref{eq:dice-token-score}.
The slope $\alpha$ corrects a drafter whose logits are too sharp or
too flat relative to the target, and $\sigma$ measures how far the
target deviates from the rescaled drafter.
As $\sigma\to0$, the estimate concentrates on
$\arg\max_v\{\alpha d_x(v)+Tg_{|x|+1}(v)\}$, and as $\sigma$ grows it
approaches the uniform distribution.
Setting $\alpha=\sigma=1$ gives a parameter-free default.

\paragraph{Path scores and node selection.}
Verification visits a node $x=x_1\cdots x_\ell$ only if the target
selects $x_j$ at the parent of every $x_j$.
If the errors are independent across nodes, the probability of this
event given the drafts and the draws is
$\prod_{j=1}^{\ell}\widehat r_{x_{<j}}(x_j)=\widehat w(x)$,
which is Eq.~\eqref{eq:dice-path-score}.
Each visited non-root node contributes one accepted token, so for a
tree $\mathcal S$ the expected number of accepted tokens is
\begin{equation}
\mathbb E\bigl[A\mid \{g_i\}\bigr]
=\sum_{x\in\mathcal S}\Pr\bigl(x\text{ is visited}\mid\{g_i\}\bigr),
\end{equation}
which DICE estimates by $\sum_{x\in\mathcal S}\widehat w(x)$.
Under a budget of $N$ nodes, this estimate is maximized by the $N$
nodes with the largest path scores.
Since $\widehat w(xv)\le\widehat w(x)$, every ancestor of a selected
node scores at least as high as the node itself, so the selected set
(with ties broken toward shallower nodes) is a valid tree.

\paragraph{Calibration.}
Every node of a verified tree reveals the target choice $y_x$,
because the target evaluates all nodes in one forward pass.
We fit $(\alpha,\sigma)$ by maximum likelihood on these choices.
With $a=\alpha/\sigma$ and $b=1/\sigma$, the log-likelihood of one
node is
\begin{equation}
\log\widehat r_x(y_x)
=a\,d_x(y_x)+bT g_{|x|+1}(y_x)
-\log\sum_v\exp\bigl(a\,d_x(v)+bTg_{|x|+1}(v)\bigr),
\end{equation}
which is concave in $(a,b)$.
The fit therefore has a unique optimum and converges in a few
gradient steps, which can also run online during decoding.

\begin{figure}[h]
  \centering
  \includegraphics[width=\textwidth]{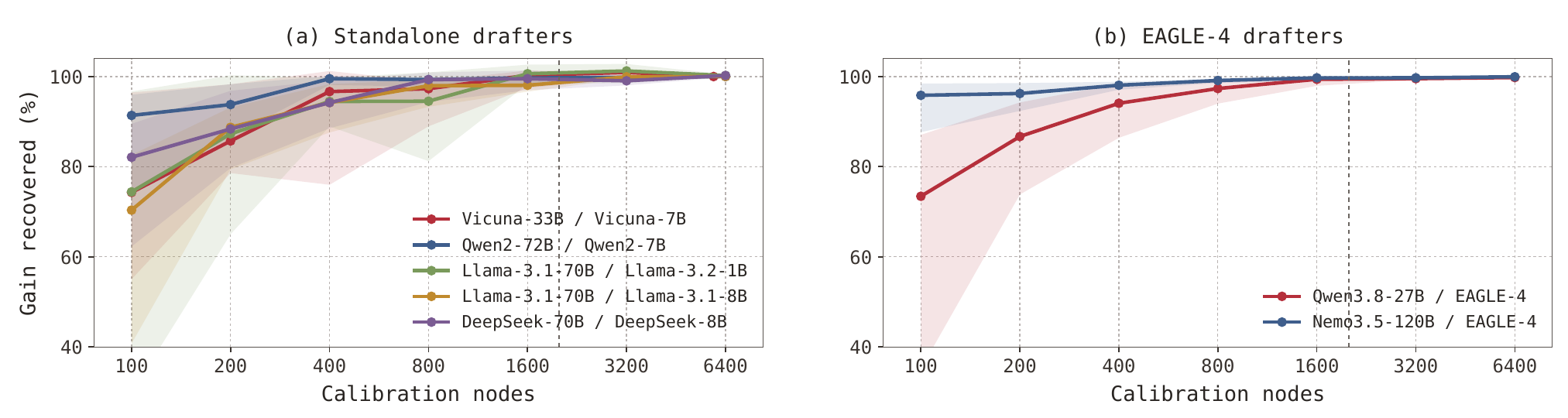}
  \caption{Share of the held-out likelihood gain recovered by fitting
  $(\alpha,\sigma)$ on a random subset of observed nodes, relative to a
  fit on all calibration data.
  Lines show the median over 20 subsets and bands the interquartile
  range. The dashed line marks the 2,000 nodes used in our experiments.}
  \label{fig:app-calib}
\end{figure}

\paragraph{How much data calibration needs.}
Figure~\ref{fig:app-calib} fits $(\alpha,\sigma)$ on random subsets
of the observed nodes for seven target--drafter pairs from
Table~\ref{tab:tree-results}, and reports the share of the held-out
likelihood gain of a fit on all calibration data that each subset
recovers.
A few hundred nodes recover most of the gain.
With 1,600 nodes, the median share is 98--100\% for every pair, and
nine in ten subsets recover at least 95\%.
The 2,000 nodes used in our experiments correspond to one to three
prompts.

\section{Correctness}
\label{app:correctness}

Throughout, the draws are indexed by absolute output position.
Let $g_1,g_2,\ldots$ be independent standard Gumbel vectors, and define
the reference sequence $y^\star$ by autoregressive Gumbel-max sampling
of the target,
\begin{equation}
y^\star_j=\arg\max_v\bigl\{z_{y^\star_{<j}}(v)+Tg_j(v)\bigr\},
\qquad j=1,2,\ldots
\end{equation}
By the Gumbel-max trick, $y^\star_j$ given $y^\star_{<j}$ is a sample
from $\operatorname{softmax}(z_{y^\star_{<j}}/T)$, so $y^\star$ follows
the target's autoregressive sampling distribution.
In the notation of Section~\ref{sec:v3-tree}, a node $x$ of a round
that starts after $j-1$ emitted tokens uses the draw
$g_{|x|+1}$ of the round, which is $g_{j+|x|}$ in absolute indexing.

\begin{proposition}[Draft-tree construction]
\label{prop:tree-exact}
For every realization of the draws and for any rule that builds each
draft tree from the prefix, the drafter, and the draws, DICE emits
exactly $y^\star$.
\end{proposition}

\begin{proof}
It suffices to show that a round starting after the prefix
$y^\star_{<j}$ emits $y^\star_j,\ldots,y^\star_{j+M}$ for some
$M\ge0$, since the next round then starts after $y^\star_{<j+M+1}$ and
the claim follows by induction over rounds.
Within the round, suppose verification has reached the node
$y^\star_j\cdots y^\star_{j+m-1}$ for some $m\ge0$, where $m=0$ is the
root.
Its prefix is $y^\star_{<j+m}$ and its draw is $g_{j+m}$, so the
target selects
$\arg\max_v\{z_{y^\star_{<j+m}}(v)+Tg_{j+m}(v)\}=y^\star_{j+m}$.
If this token is a child of the node, verification accepts it and
moves to the child.
Otherwise it emits $y^\star_{j+m}$ as the bonus token and the round
ends.
Either way, the round emits the next token of $y^\star$, and since the
tree is finite, the round ends after emitting
$y^\star_j,\ldots,y^\star_{j+M}$.
\end{proof}

The draft tree may depend on the draws in any way, but it only
determines how many tokens each round emits, not which tokens.
This is why conditioning tree construction on the draws leaves the
output distribution unchanged.

\begin{proposition}[Asynchronous prefetching]
\label{prop:prefetch-exact}
SSD with the DICE prefetch policy has the same output distribution as
the target.
\end{proposition}

\begin{proof}
Each round of SSD verifies a proposal $x_{1:K}$ by speculative
sampling.
It accepts $x_i$ if $u_iq_i(x_i)\le p_i(x_i)$, and samples the
correction or bonus token from the normalized residual
$[p_{k+1}-q_{k+1}]_+$ or from $p_{K+1}$ by the Gumbel-max trick with
$g_{k+1}$.
Speculative sampling yields the target distribution as long as, given
the prefix, (i) the proposal is drawn from the draft distributions and
(ii) the verification draws are independent of the proposal
\citep{leviathan2023,chen2023}.
Both conditions hold under DICE.
The drafter samples every proposal from $q_i$ with its own randomness,
and the prefetch policy only chooses which prefixes to draft in
advance, so the proposal for the realized prefix is a draw from $q_i$
whether it was prefetched or drafted after a cache miss.
The verification draws are sampled independently of the drafter and
are never used to generate proposals.
The policy reads them only to decide which drafts to prepare, which
changes when a draft becomes available but not its value.
Every round is therefore a standard speculative-sampling round, and
the output follows the target distribution.
\end{proof}

\section{Maximum-Entropy Ratio Model}
\label{app:maxent}

\paragraph{Derivation.}
For a proposed token $x$ with draft probability $q=q_i(x)$, the ratio
$\rho=p_i(x)/q$ lies in $[0,b]$ with $b=1/q$, since $p_i(x)\le1$.
If the drafter is calibrated, $\mathbb E[p_i(x)\mid q]=q$, so
$\mathbb E[\rho\mid q]=1$.
Among densities on $[0,b]$ with mean one, the one with maximum entropy
solves
\begin{equation}
\max_f\ -\!\int_0^b f\log f\,d\rho
\quad\text{s.t.}\quad
\int_0^b f\,d\rho=1,\qquad \int_0^b\rho f\,d\rho=1.
\end{equation}
Setting the variation of the Lagrangian to zero gives
$\log f(\rho)=-1-\mu-\lambda\rho$, so
\begin{equation}
f_q(\rho)=\frac{\lambda e^{-\lambda\rho}}{1-e^{-\lambda b}},
\qquad 0\le\rho\le b,
\label{eq:app-maxent}
\end{equation}
a truncated exponential distribution, with $f_q$ uniform when
$\lambda=0$.
The rate $\lambda$ is fixed by the mean constraint,
\begin{equation}
\frac1\lambda-\frac{b}{e^{\lambda b}-1}=1.
\label{eq:app-lambda}
\end{equation}
The mean of $f_q$ decreases strictly in $\lambda$, from $b$ as
$\lambda\to-\infty$ to $0$ as $\lambda\to\infty$, and equals $b/2$ at
$\lambda=0$.
Since $0<1<b$, Eq.~\eqref{eq:app-lambda} has a unique solution, which
depends on $q$ alone.
It is positive when $q<1/2$, zero when $q=1/2$, and negative when
$q>1/2$.
We tabulate $\lambda$ over $q$ once, so the model has no fitted
parameters.

\paragraph{Acceptance probability.}
A proposal is accepted if and only if $\rho\ge u_i$.
Given the known draw $u_i$, Eq.~\eqref{eq:dice-accept-prob} has the
closed form
\begin{equation}
\widehat\eta_i=\Pr_{\rho\sim f_q}(\rho\ge u_i)
=\frac{e^{-\lambda u_i}-e^{-\lambda b}}{1-e^{-\lambda b}}
\quad(\lambda\ne0),
\qquad
\widehat\eta_i=1-\frac{u_i}{b}\quad(\lambda=0).
\end{equation}
Averaged over $u_i\sim\mathrm{Uniform}(0,1)$, this gives
$\mathbb E[\min(\rho,1)]$, the acceptance probability of speculative
sampling under the model.

\paragraph{Correction and bonus tokens.}
Under the same model, $q_{k+1}(v)\rho_{k+1}(v)$ plays the role of the
target probability $p_{k+1}(v)$.
The weights in Eq.~\eqref{eq:dice-prefetch-weights} are therefore the
residual $[p_{k+1}-q_{k+1}]_+$ after a rejection and the target
distribution $p_{K+1}$ after full acceptance, which are the two
distributions that verification samples from.
Holding $g_{k+1}$ fixed and maximizing $\log m_k(v)+g_{k+1}(v)$ over
sampled ratios reproduces the Gumbel-max step of verification, so the
resulting frequencies estimate the probability of each token given the
known draw.
A rejected proposal satisfies $\rho_{k+1}(x_{k+1})<u_{k+1}\le1$, so its
residual weight is zero, as in verification.

\section{Seed-Aware Asynchronous Prefetching}
\label{app:prefetch}

Algorithm~\ref{alg:prefetch} summarizes the procedure of
Section~\ref{sec:v3-prefetch}.
In our implementation, each row keeps $M=16$ candidates and uses
$S=128$ fixed stratified samples of $\rho$.
Samples in which $m_k(v)=0$ for every candidate are discarded;
if no sample remains, $\widehat r_k$ falls back to
$\operatorname{softmax}(\log q_{k+1}+g_{k+1})$ over $\mathcal V_k$.
The cache budget $B$ equals the default cache size of SSD.

\begin{algorithm}[h]
\caption{DICE for Asynchronous Prefetching}
\label{alg:prefetch}
\begin{algorithmic}[1]
\REQUIRE Prefix $h$, proposal $x_{1:K}$,
draft distributions $\{q_i\}_{i=1}^{K+1}$,
draws $\{u_i\}$ and $\{g_i\}$, cache budget $B$,
candidates $M$, samples $S$
\ENSURE Prefetch cache $\mathcal K$, $|\mathcal K|\le B$
\dicestage{Estimate stopping probabilities}
\STATE $\widehat\eta_i\gets\Pr_{\rho\sim f_{q_i(x_i)}}(\rho\ge u_i)$,
\quad $i=1,\ldots,K$
\STATE $\widehat P(k)$ by Eq.~\eqref{eq:dice-stop-prob},
\quad $k=0,\ldots,K$
\dicestage{Estimate correction and bonus tokens}
\FOR{$k=0,\ldots,K$}
  \STATE $\mathcal V_k\gets\operatorname{Top}_M(\log q_{k+1}+g_{k+1})
  \setminus\{x_{k+1}\}$
  \STATE $\rho^{(s)}(v)\sim f_{q_{k+1}(v)}$,
  \quad $v\in\mathcal V_k$, $s=1,\ldots,S$
  \STATE $v^{(s)}\gets\arg\max_{v\in\mathcal V_k}
  \big\{\log m_k^{(s)}(v)+g_{k+1}(v)\big\}$
  \quad (Eq.~\eqref{eq:dice-prefetch-weights})
  \STATE $\widehat r_k(v)\gets$ share of valid samples with $v^{(s)}=v$
  \STATE $\widehat\pi(k,v)\gets\widehat P(k)\,\widehat r_k(v)$
\ENDFOR
\dicestage{Allocate the prefetch cache}
\STATE $\mathcal C\gets\operatorname{Top}_B(\{(k,v)\}_{k,v},\widehat\pi)$
\FOR{$(k,v)\in\mathcal C$}
  \STATE $\mathcal K[hx_{1:k}v]\gets
  \textsc{Prefetch}_{\mathrm{SSD}}(hx_{1:k}v)$
\ENDFOR
\STATE \textbf{return} $\mathcal K$
\end{algorithmic}
\end{algorithm}

\end{document}